\documentclass[10pt,twocolumn,letterpaper]{article}

\usepackage{wacv}
\usepackage{times}
\usepackage{epsfig}
\usepackage{graphicx}
\usepackage{amsmath}
\usepackage{amssymb}
\usepackage{authblk}
\usepackage{placeins}
\usepackage[symbol*]{footmisc}

\makeatletter
\newcommand{\printfnsymbol}[1]{%
  \textsuperscript{\@fnsymbol{#1}}%
}
\makeatother

\usepackage{graphicx}
\usepackage[ruled,vlined]{algorithm2e}
\usepackage{multirow}
\usepackage{float}
\usepackage{caption}
\usepackage{paralist}

\def\wacvPaperID{917} 

\wacvfinalcopy 
\ifwacvfinal
\def\assignedStartPage{1} 
\fi

\usepackage{hyperref}
\hypersetup{breaklinks=true,linkcolor=blue,citecolor=blue}

\ifwacvfinal
\else
\fi

\begin{document}

\title{Multi-path Neural Networks for On-device Multi-domain Visual Classification}

\author[1]{Qifei Wang \thanks{equal contribution}}
\author[1]{Junjie Ke \printfnsymbol{1}}
\author[2]{Joshua Greaves}
\author[1]{Grace Chu}
\author[2]{Gabriel Bender}
\author[1]{Luciano Sbaiz}
\author[1]{Alec Go}
\author[1]{Andrew Howard}
\author[1]{Ming-Hsuan Yang}
\author[1]{Jeff Gilbert}
\author[1]{Peyman Milanfar}
\author[1]{Feng Yang}
\affil[1]{Google Research, Google Brain} 
\affil[2]{Google Brain}
\affil[ ]{\textit {\{qfwang, junjiek, joshgreaves, cxy, gbender, sbaiz, ago, howarda, minghsuan, jegilbert, milanfar, fengyang\}@google.com}}

\maketitle
\thispagestyle{empty}



\begin{abstract}
Learning multiple domains/tasks with a single model is important for improving data efficiency and lowering inference cost for numerous vision tasks, especially on resource-constrained mobile devices. 
However, hand-crafting a multi-domain/task model can be both tedious and challenging. 
This paper proposes a novel approach to automatically learn a multi-path network for multi-domain visual classification on mobile devices. 
The proposed multi-path network is learned from neural architecture search by applying one reinforcement learning controller for each domain to select the best path in the super-network created from a MobileNetV3-like search space. 
An adaptive balanced domain prioritization algorithm is proposed to balance optimizing the joint model on multiple domains simultaneously. 
The determined multi-path model selectively shares parameters across domains in shared nodes while keeping domain-specific parameters within non-shared nodes in individual domain paths. 
This approach effectively reduces the total number of parameters and FLOPS, encouraging positive knowledge transfer while mitigating negative interference across domains. 
Extensive evaluations on the Visual Decathlon dataset demonstrate that the proposed multi-path model achieves state-of-the-art performance in terms of accuracy, model size, and FLOPS 
against other approaches using MobileNetV3-like architectures.
Furthermore, the proposed method improves average accuracy over learning single-domain models individually, and reduces the total number of parameters and FLOPS by $78\%$ and $32\%$ respectively, compared to the approach that simply bundles single-domain models for multi-domain learning.
\end{abstract}

\begin{figure*}
\centering
    \includegraphics[trim=2cm 5cm 1.5cm 4cm, clip=true, width=125mm,scale=0.6]{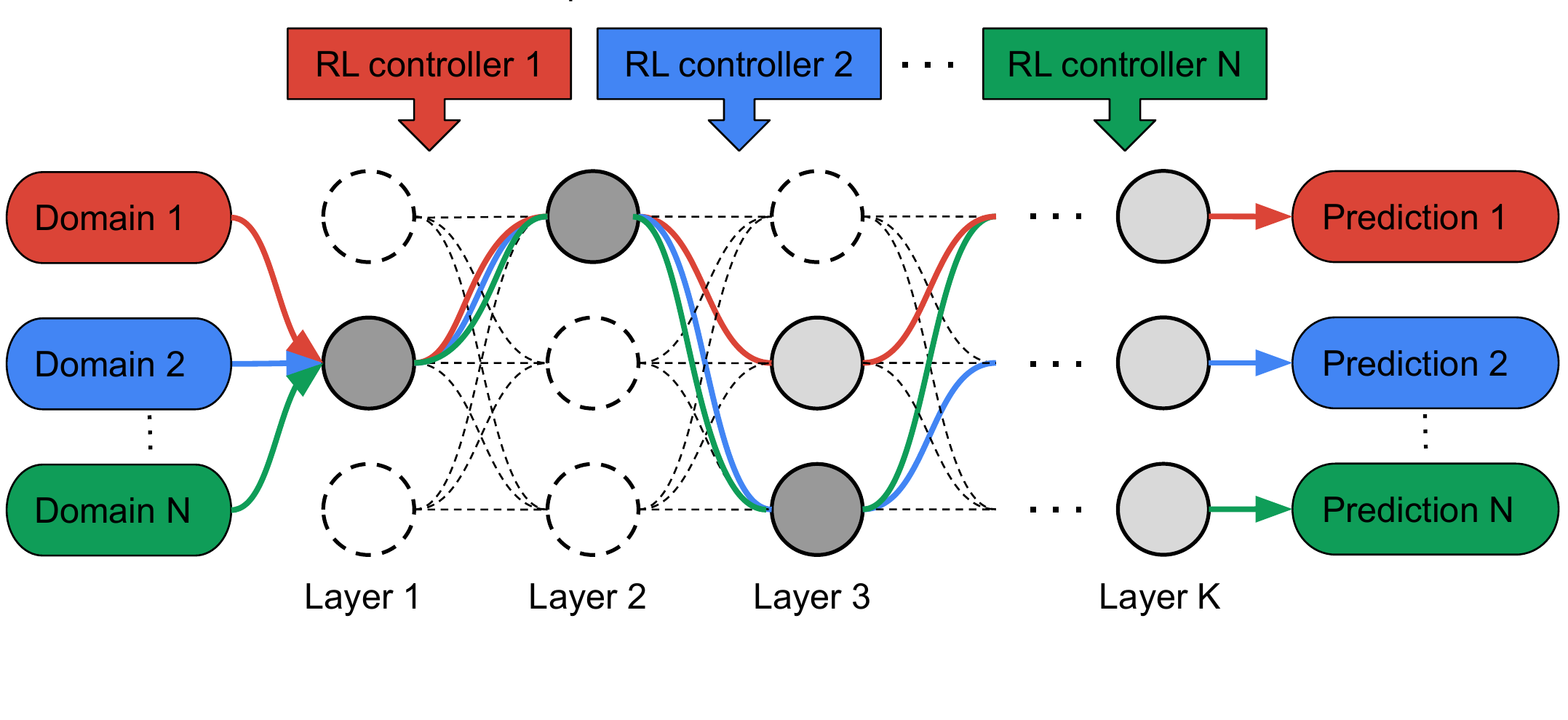}
    \setlength{\abovecaptionskip}{0.cm}
    \setlength{\belowcaptionskip}{-0.cm}
    \caption{Multi-path network for multi-domain visual classification. Different color paths show model architectures for different domains. For example, the red path represents the model architecture formed for domain 1. Paths that run through the same node end up sharing the weights of the shared node (dark grey).}
    \label{fig:multi-route-example}
\end{figure*}

\section{Introduction}

Numerous methods based on deep learning have made 
significant advances in a wide range of vision tasks, including image classification~\cite{krizhevsky2012imagenet}, object detection~\cite{girshick2014rich}, and segmentation~\cite{he2017mask}, to name a few. 
However, most deep neural networks (DNNs) are developed for a single task with data coming from the same domain. 
There is growing interest in training a joint model to tackle multiple domains and tasks, i.e., multi-domain learning (MDL) as well as multi-task learning (MTL).

MDL and MTL are two overlapping but distinct problems.
MTL~\cite{vandenhende2020revisiting} focuses on performing multiple related tasks, e.g., segmentation and depth estimation on a given data sample. 
On the other hand, MDL aims to build a joint model to perform the same task, e.g., classification, across multiple visual domains, such as Internet images, characters, glyph, animal, sketches, etc. 
Building a joint MDL model is especially important on resource-constrained mobile devices, where deploying a separate model for each domain can introduce high latency, large memory footprint and power consumption. 
Although numerous mobile friendly models (e.g., MobileNetV3~\cite{howard2019searching}, ShuffleNet~\cite{zhang2018shufflenet}, MNAS~\cite{tan2019mnasnet}, and ENAS~\cite{pham2018efficient}) have been proposed for single domain learning, much less attention has been paid to develop efficient MDL systems for mobile devices.

One straightforward MDL approach is to train a universal feature extractor for all domains. 
However, its performance may degrade significantly due to the negative transfer between different domains. 
Existing MDL approaches~\cite{rebuffi2017learning, rebuffi2018efficient} start from a pre-trained model and  fine-tune with added domain specific modules. 
Each domain can therefore selectively use the features from the fixed main backbone network. 
Although this approach naturally reduces the \emph{negative transfer} effect between domains, it does not encourage \emph{positive transfer} between domains due to the frozen backbone network. 
%
%
Manually designing domain-adaptive modules can also be challenging and tedious since it is difficult to transfer the manual modules onto different architectures to achieve both high classification accuracy across domains and low costs in terms of computation and memory consumption.

Another challenging factor for learning a joint MDL model is domain prioritization during training. 
Different domains usually have diverse degrees of difficulty which may not be constant during the training process. 
Although various optimization-based approaches~\cite{kendall2018multi, chen2018gradnorm, liu2019end, guo2018dynamic, sener2018multi} have been proposed to deal with this issue, jointly balancing domain difficulties to achieve optimal performance across multiple domains remains a difficult problem.

In this paper, we show that a good MDL model for mobile devices should: 1) achieve high accuracy while keeping number of parameters and FLOPS low; 2) learn to enhance positive knowledge sharing while mitigating negative transfer among domains; and 3) effectively optimize the joint model across domains and be adaptive to domain difficulties. 
As such, we propose a neural architecture search (NAS) method  for designing effective MDL models, and train it with adaptive balanced domain prioritization. 
A multi-path NAS approach is developed to automatically design an MDL model with a learned parameter sharing strategy among domains. 
Based on the single-domain efficient NAS framework~\cite{bender2020can}, the proposed multi-path NAS for MDL uses multiple reinforcement learning (RL) controllers, where each selects an optimal path from the super-network for each domain. 
To ensure mobile efficiency, the multi-path NAS network is learned based on the MobileNetV3-like search space and each RL controller optimizes its domain path based on the trade-off between accuracy and inference cost.
Fig.~\ref{fig:multi-route-example} shows the multi-path architecture search framework. The model for one domain is represented by the nodes wired by a single color path.

An adaptive balanced domain prioritization algorithm is further proposed to balance the learning progress dynamically for domains of diverse difficulties in both the architecture search and model training phases. 
The selected multi-path model demonstrates consistent advantages when evaluated on the Visual Decathlon dataset \cite{rebuffi2017learning}, reducing the model size and FLOPS by $78\%$ and $32\%$ without sacrificing the average accuracy compared to the approach that simply constructs single-domain models.

The main contributions of this work are:
\begin{compactitem}
\item We propose a generic multi-path neural architecture search (MPNAS) approach as the first work to apply ENAS
to solve the challenges of on-device MDL, including
data imbalance, domain diversity, negative transfer, domain scalability, and large search space of possible parameter sharing strategies. 
The MPNAS approach learns a cross-domain parameter sharing strategy to encourage positive knowledge transfer and also mitigate negative knowledge transfer via domain-wise path selection.
The MPNAS approach achieves state-of-the-art accuracy, model size, and FLOPS on MobileNetV3-like architectures. 
\item We improve MDL model optimization on domains of diverse difficulties by using the generic adaptive balanced domain prioritization in the objective function. 
\item We present comprehensive studies for detailed insights into the network architecture of the learned MobileNetV3-like MDL model and the effect of different domain weighting approaches for MDL.
\end{compactitem}

\begin{figure*}[tp]
\centering
\includegraphics[trim=0.0cm 9.5cm 0.0cm 1cm, clip=true, width=150mm,scale=1]{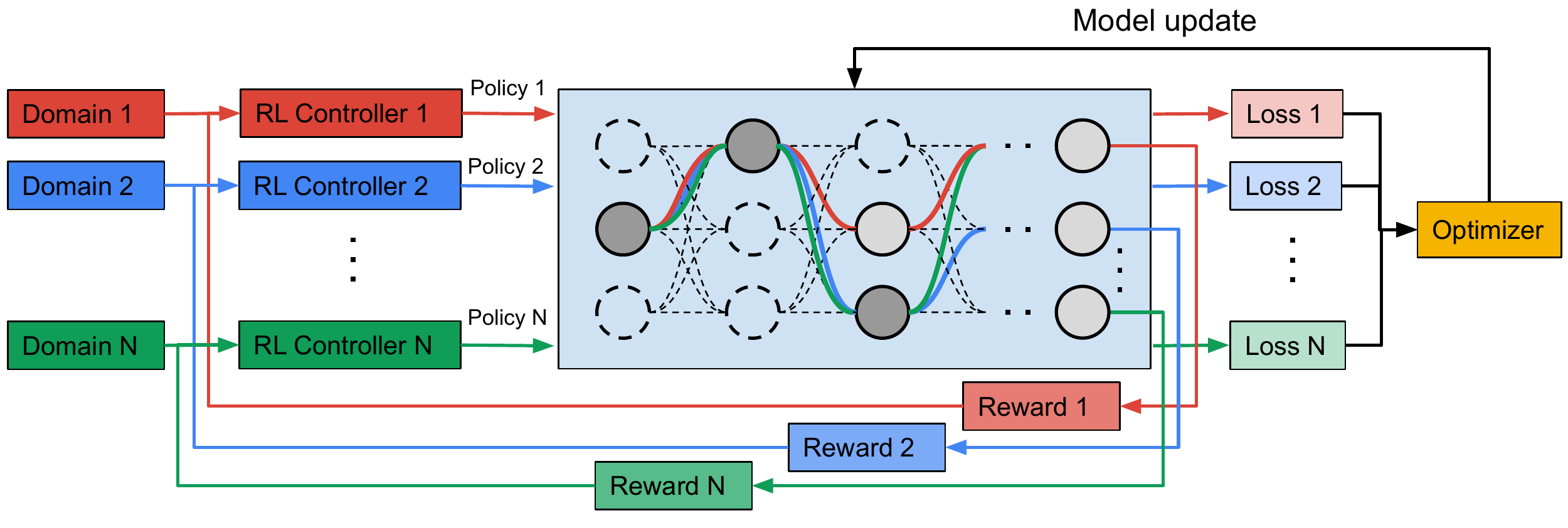}
\setlength{\abovecaptionskip}{0.cm}
\setlength{\belowcaptionskip}{-0.cm}
\caption{Multi-path neural architecture search framework. Each circle represents a candidate node in the super-network. Dark nodes are shared among domains. Light nodes are used by a single domain (non-shared). Dashed nodes are not selected by any domain and will be removed from the final model.}
\label{fig:multi-path-nas}
\end{figure*}

\section{Related Work}

\vspace{2mm}
\noindent \textbf{Multi-Domain Learning.} MDL focuses on learning a joint deep neural network that performs well across multiple input domains. 
Most recent methods adapt pre-trained models to new tasks by adding few task-specific parameters. 
Bilen and Vedaldi~\cite{bilen2017universal} proposed a single network to learn the universal features for all domains by sharing all the parameters in the feature extraction layers except batch and instance normalization layers. 
Rebuffi et al.~\cite{rebuffi2017learning} extended the universal feature extraction network by introducing an adapter residual module to the standard residual network which makes the network adaptive to different visual domains. 
Further extensive studies~\cite{rebuffi2018efficient} have been made on series and parallel residual adapters, joint adapter compression, and parameter allocations, which results in an empirically optimized network for MDL. 
Berriel et al. presented a budget-aware adapter module \cite{berriel2019budget} to select the most relevant feature channels for each domain. 
On the other hand, incremental learning~\cite{rosenfeld2018incremental} is adopted for learning new domains sequentially without forgetting the knowledge learned on the previous domains. 
Existing MDL approaches showed that positive knowledge transfer can improve the performance over single domain learning with limited increase of model size.
However, the hand-crafted sharing schemes and empirical optimization approaches which are designed for certain types of networks such as residual networks may not be generally applicable to other network architectures, e.g., mobile friendly architectures such as MobileNet~\cite{howard2019searching}.

\vspace{2mm}
\noindent \textbf{Multi-Task Learning.} MTL aims to simultaneously learn a diverse set of tasks, e.g., object detection, segmentation, depth estimation, by sharing knowledge among them. 
Different MTL models differ in strategies of sharing filters \cite{newell2019feature,sun2019adashare,strezoski2019many,mallya2018piggyback} or sharing operations \cite{lu2017fully, vandenhende2019branched, guo2020branch} among tasks, aiming to improve generalization by mining task relations and to suppress negative transfer. 
In contrast to our work, this line of research focuses on learning a diverse set of tasks in the same visual domain, usually all operating on the same image.

Other than hand-crafted MTL networks, routing networks \cite{rosenbaum2017routing} use an RL agent to learn the best route layer-by-layer. However, this method scales poorly with the size of the search space. 
A multi-task architecture search (MAS) \cite{pasunuru2019continual} is proposed to search for a universal feature extraction architecture, but suffers from  negative transfer among tasks.

\vspace{2mm}
\noindent \textbf{Domain/Task Balanced Optimization.} Since different domains/tasks have different levels of difficulties, one of the main challenges in training an MDL/MTL model is to strike a balance between domains/tasks when training simultaneously. 
Task balancing has been studied extensively in the MTL literature. 
Kendall et al. \cite{kendall2018multi} used homoscedastic uncertainty as the weight which favors the task with less noisy annotations. 
Dynamic weight averaging (DWA) \cite{liu2019end} is proposed to balance the learning pace across tasks to optimize all the tasks at a similar pace. 
Neither approach takes into consideration the task difficulties which may vary greatly in applications. 
To address this issue, the Dynamic task prioritization (DTP) \cite{guo2018dynamic} method is developed to prioritize difficult tasks by adjusting the weight of each single-domain loss dynamically. 
However, the DTP method needs a surrogate measurement for task difficulty, which may be impractical for certain problems. 
To be agnostic to the task difficulties, the balanced multi-task learning loss (BMTL) function \cite{liang2020simple} is proposed and shown to achieve promising results on MTL. 
%
Similar to the BMTL scheme, we develop a domain/task balancing method for MDL.
In MDL, loss functions for different domains often share the same type and are of the same magnitude, which means loss function can be a good surrogate for task difficulties. 
However, this can be unrealistic in MTL (e.g., classification and segmentation tasks) where loss types differ. 
More comprehensive overview on MTL optimization can be found in~\cite{vandenhende2020revisiting}.

\vspace{2mm}
\noindent \textbf{Neural Architecture Search.} NAS is a powerful paradigm for automatically designing neural architectures. 
Recent NAS work shift the focus from the expensive evolutionary NAS to efficient NAS \cite{pham2018efficient, liu2018darts, xie2018snas, cai2018proxylessnas} by constructing a super-network graph based on the search space and learning the architecture via end-to-end path sampling. 
While the existing efficient NAS frameworks achieve significant success on searching architectures for a single domain, 
finding effective MDL models still remains an active area. 
We propose a multi-path NAS method for MDL based on the latest single-domain NAS work TuNAS \cite{bender2020can} which provides the advantages of handling substantially large and difficult search spaces to build mobile-friendly models without domain specific prior knowledge.

\section{Multi-path NAS for MDL}

The main modules of the proposed multi-path NAS algorithm for MDL are shown in Fig.~\ref{fig:multi-path-nas}.
At the search stage, a super-network with multiple candidate paths (both solid and dot paths in Fig.~\ref{fig:multi-path-nas}) is constructed. 
Assuming that there are $N$ domains $D = \{D_1, D_2, \dots, D_N\}$, the proposed multi-path network uses $N$ RL controllers $C = \{C_1, C_2, \dots, C_N\}$, to manage the path selection for each domain.
The circles represent candidate nodes in the super-network which are defined by the search space and the RL controllers learn to select nodes to build paths for each domain. 
The final model is formed by merging all the selected paths into a single network.

At each search iteration, $C_i$ samples a single path for domain $D_i$. 
Each sampled path forms a sampled model for domain $D_i$. 
We first update the model weights using a single batch of images from each domain on the training set, then update the weights of RL controllers using the validation set for the corresponding domain. 
Specifically, the RL controller $C_i$ receives a reward $\mathcal{R}_i$ 
(see section \ref{section:reward-function} for a function of validation accuracy and latency) from the model prediction. 
This $\mathcal{R}_i$ is then used to perform policy gradient update (REINFORCE \cite{williams1992simple}) on $C_i$. 
The process is repeated until it reaches the maximum epochs. 
The main search steps for the multi-path network are summarized in Algorithm \ref{algorithm1}.

At the end of the search phase, we take the most likely paths from each RL controller and form a single joint model with all the paths. 
As shown in Fig.~\ref{fig:multi-path-nas}, the model for one domain is constructed with the nodes connected by a path in one color. 
If more than one domain is connected to the same node in the super-network, the weights in that shared node will be used by all those domains (dark nodes with solid circles). 
If only one domain is routed to a node, the node will be exclusively used by that domain (light nodes with solid circles).  
Inside a conv node, each domain can selectively use a subset of filters. 
The filter number is also selected by the RL controller. 
The nodes not used by any domain will be removed from the final model (nodes with dashed circles). 

The joint model is then trained from scratch with training data from all domains. 
We obtain domain predictions by running the corresponding domain data through its own path in the joint model.  
The final training process is similar to Algorithm \ref{algorithm1} but with a fixed model and no RL controller. 
In each batch we compute each domain's training loss TrainLoss[i] by running the fixed model on domain training set. The joint TrainLoss, which is the sum of each individual domain's TrainLoss[i], will be back-propagated to update the model weights . 
The gradients for parameters in those shared nodes will be updated jointly.

\begin{algorithm}[tp]
 \caption{Multi-path neural architecture search}
 \label{algorithm1}
\SetAlgoLined
\KwResult{Multi-path network}
 Initialize RLControllers\;
 Initialize super-network from the search space\;
 \For{$Epochs = 1 : MaxEpochs$}{
  \For{$i = 1 : DomainCount$}{
    Sample one path for Domain[i] to form model\;
    Run model on validation set to get Reward[i]\;
    Run model on training set to get TrainLoss[i]\;
  }
  
  Backprop the joint TrainLoss to update model coefficients in super-network\;
  \For{$i = 1 : DomainCount$}{
    Update RLControllers[i] with Reward[i] using REINFORCE\;
  }
 }
\end{algorithm}

\begin{figure*}[htbp]
\centering
\includegraphics[width=173mm]{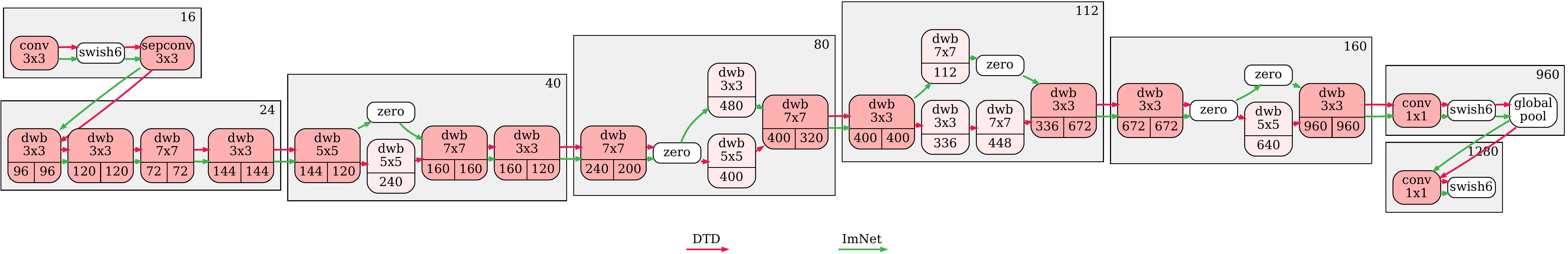}
\setlength{\abovecaptionskip}{0.cm}
\setlength{\belowcaptionskip}{-0.cm}
\caption{Multi-path model for ImageNet72 (green path) and Describable Textures (red path). Darker nodes are shared between domains while lighter nodes are independent. The numbers under the nodes denote the number of filters used by each domain in that conv node.}
\label{fig:two-path-model}
\end{figure*}

Fig.~\ref{fig:two-path-model} shows a partial multi-path model architecture for the two domains on the Visual Decathlon dataset. 
Given that each RL controller independently selects the path based on the domain accuracy reward, the path similarity can also manifest the correlations among domains (see Section~\ref{section:architecture-analysis} for domain correlation analysis).  

\subsection{Reward Function}
\label{section:reward-function}
In order to determine architectures with good accuracy and latency trade-offs, we use the parameterized RL reward function proposed in the MnasNet \cite{tan2019mnasnet} (also adopted by the ProxylessNAS \cite{cai2018proxylessnas}):
\begin{align}
\label{eq:reward-function}
   r(\alpha) = Q(\alpha) \times (T(\alpha) / T_0)^\beta,
\end{align}
where $Q(\alpha)$ represents the quality (validation accuracy) of a candidate architecture $\alpha$, $T(\alpha)$ is the estimated inference time and $T_0$ is the target inference time. 
The cost exponent $\beta < 0$ is a tunable hyper-parameter. The reward function is maximized when model quality $Q(\alpha)$ is high and inference time $T(\alpha)$ is small.

\section{Domain Prioritization in Multi-path NAS}

Handling domains with diverse difficulties is one of the main challenges in MDL.
Given each domain has a different number of classes and difficulties, a method that directly optimizes the sum of losses is unlikely to perform well as shown in the ablation studies (Section~\ref{sec:ablation_studies:ABDP}). 
%
%
We propose an adaptive balanced domain prioritization (ABDP) method to achieve balanced training performance for all the domains by introducing a transformed loss objective function:
\begin{align}
\label{eq:bmtl}
    \min_{\theta} \sum\limits_{i=1}^N h(\mathcal{L}(D_i; \theta)) + r(\theta),
\end{align}
where $\theta$ denotes the model parameters, $\mathcal{L}(D_i; \theta)$ represents the loss on domain $D_i$ with the current model parameters, and $r(\theta)$ is the regularization term. 

Since all domains share the same loss function in MDL, the loss $\mathcal{L}(D_i; \theta)$ can be a good surrogate to indicate the difficult of domain $D_i$.
The boosting function $h(\cdot)$ in equation (\ref{eq:bmtl}) is introduced to transform the loss subspace to a new subspace to boost the priorities of more difficult domains. 
Its derivative $h'(\mathcal{L}(D_i; \theta))$ can be viewed as the weight of the current domain $D_i$ during gradient update. 
When $h(\cdot)$ is monotonically increasing, the domains with larger losses are favored.  
The loss function in Eq.~(\ref{eq:bmtl}) is adaptive in nature since it can dynamically adjust domain weights during the entire training phase.

If $h(\cdot)$ is a linear function, the objective function regresses to a linear weight of the single-domain losses which cannot achieve desired domain prioritization since $h'(\cdot)$ is constant. 
In the ablation studies (see Section~\ref{sec:ablation_studies:ABDP}), we evaluate multiple options for the boosting function $h(\cdot)$, including linear, polynomial, and exponential functions. 
Both the polynomial and exponential functions can amplify the loss which means the optimizer favours more difficult domains over easier ones. 
Our ablation studies show that a nonlinear boosting function can improve the model performance in MDL. Empirically, the exponential boosting function can most effectively boost performance on hard domains.

We further make the joint loss function adjustable during the search and training process by introducing a domain prioritization coefficient $w$ in the boosting function. 
Assuming that we use the exponential function as the boosting function, the adaptive boosting function can be defined as:
\begin{align}
\label{eq:adaptive-bmtl}
    h = \exp\Big(\frac{\mathcal{L}(D_i; \theta)}{w}\Big).
\end{align}
In Eq.~(\ref{eq:adaptive-bmtl}), the adaptive parameter $w$ can be put on a decay schedule throughout the training phase (e.g. linear decay from $w_{max}$ to $w_{min}$). 
As $w$ decreases, the domains with larger loss will become increasingly more important, which means the model favors difficult domains more at the later search/training stage. 
From our ablation studies, decreasing schedule of $w$ outperforms constant schedule on the Visual Decathlon dataset. 
The results verify that adaptive prioritization can further optimize the MDL performance.

\section{Experiments}

We evaluate the proposed mobile-friendly MDL models in three aspects.
First, we evaluate the number of parameters, FLOPS, and Top-1 accuracy of the MDL model constructed by the proposed multi-path NAS (MPNAS) model vs. other state-of-the-art approaches based on MobileNetV3-like architectures, including single domain NAS model bunlding, single path MobileNet pre-searched model, multi-domain single path NAS model, multiple random path model, and task routing layer (TRL) model proposed in~\cite{strezoski2019many}.
Second, we analyze the selected MPNAS model and evaluate domain similarity by calculating the Jaccard similarity score~\cite{gower2014similarity} between domain path pairs.
Third, we carry out ablation studies of the adaptive balanced domain prioritization.

\vspace{1mm}
\noindent \textbf{Datasets.} We carry out experiments on the Visual Decathlon dataset \cite{rebuffi2017learning}, which is composed of ten widely-used image classification problems representing sufficiently diverse visual domains. 
Each domain of this dataset has diverse image/class sizes and difficulties, which is crucial to verify the performance of MDL.
The image resolution of the ten domains are normalized to $72 \times 72$ pixels which makes the classification task more challenging.

\setlength{\tabcolsep}{3pt}
\begin{table*}[tp]
\begin{center}
\caption{Number of parameters, FLOPS(G), and Top-1 accuracy (\%) of MDL models on the Visual Decathlon dataset. All the methods are built based on the MobileNetV3-like search space. Blue number indicates the best result.}
\label{table:multihead-fulltrain}
\begin{tabular}{c|c|c|cccccccccc|c}
\hline\noalign{\smallskip}
\multirow{2}{*}{Models} & \# & FLOPS & \multirow{2}{*}{ImageNet72} & \multirow{2}{*}{Airc.} & \multirow{2}{*}{C100} & \multirow{2}{*}{DPed} & \multirow{2}{*}{DTD} & \multirow{2}{*}{GTSR} & \multirow{2}{*}{Flwr} & \multirow{2}{*}{OGlt} & \multirow{2}{*}{SVHN} & \multirow{2}{*}{UCF} & \multirow{2}{*}{Mean} \\
& par. & (G) &  &  &  &  &  &  &  &  &  &  &  \\
\noalign{\smallskip}
\hline
\noalign{\smallskip}
SD-NAS & 5.7x & 1.08 & \textbf{\textcolor{blue}{61.9}} & 22.6 & 69.6 & \textbf{\textcolor{blue}{98.4}} & 34.2 & 99.5 & 66.7 & \textbf{\textcolor{blue}{78.7}} & \textbf{\textcolor{blue}{93.3}} & \textbf{\textcolor{blue}{74.4}} & 69.9 \\
SP-MN & 1.0x & 0.09 & 30.2 & 6.2 & 59.5 & 59.9 & 31.0 & 52.5 & 60.2 & 1.3 & 23.1 & 28.8 & 35.2 \\
MDSP-NAS & \textbf{\textcolor{blue}{0.7x}} & \textbf{\textcolor{blue}{0.04}} & 34.6 & 18.8 & 56.4 & 75.1 & 30.8 & 88.6 & 65.6 & 9.3 & 25.5 & 47.5 & 45.2 \\
MRP & 1.3x & 0.43 & 50.7 & 21.5 & 67.3 & 96.1 & 31.5 & \textbf{\textcolor{blue}{99.8}} & 61.8 & 75.8 & 91.9 & 69.5 & 66.6 \\ 
TRL \cite{strezoski2019many} & 1.0x & 0.71 & 27.1 & 16.1 & 60.0 & 66.2 & 33.9 & 95.8 & 63.3 & 54.3 & 85.7 & 46.2 & 54.9 \\ 
MPNAS & 1.3x & 0.73 & 57.0 & \textbf{\textcolor{blue}{29.7}} & \textbf{\textcolor{blue}{81.2}} & 96.0 & \textbf{\textcolor{blue}{36.6}} & \textbf{\textcolor{blue}{99.8}} & \textbf{\textcolor{blue}{79.2}} & 74.2 & 92.3 & 71.8 & \textbf{\textcolor{blue}{71.8}} \\ 
\hline
\end{tabular}
\end{center}
\end{table*}
\setlength{\tabcolsep}{1pt}

\vspace{1mm}
\noindent \textbf{Evaluated methods.}
We evaluate the proposed method against the state-of-the-art approaches including:
%
\begin{compactitem}




\item \textnormal{Single domain NAS model (SD-NAS)}: A domain independent model selected on the MobileNetV3-like search space and trained from scratch on a single domain dataset.
\item \textnormal{Single path MobileNet pre-searched model (SP-MN)}: A model whose backbone is searched using ImageNet72 based on the MobileNetV3-like search space. Domain specific output heads are attached to the final FC layer. All the domains share the main feature extraction network body.
\item \textnormal{Multi-domain single path NAS model (MDSP-NAS)}: A single-path model selected by a single RL controller based on the MobileNetV3-like search space. 
The joint RL controller learns to select the path with the average accuracy as reward. Like SP-MN, all domains use a shared network body with each output head attached to the final FC layer.
\item \textnormal{Multiple random path model (MRP)}: A joint model with a random path assigned to each domain. 
The paths are randomly selected from the super-network generated by the MobileNetV3-like search space. 
Results are averaged across ten trials to reduce noise.

\item \textnormal{Task routing layer model (TRL)}: The model with task routing layer for MDL proposed in \cite{strezoski2019many}. For fair comparisons, we add the task routing layer on top of the MobileNetV3-Large baseline model. 
Since the capacity of MobileNetV3-like models is very small,
%
we set the ratio of active filters for each task to 0.9 which achieves the best performance in our experiments.
\item \textnormal{Multi-path NAS model (MPNAS)}: The proposed multi-path model obtained by using multiple RL controllers to select paths for each domain based on the MobileNetV3-like search space. 
The shared and domain specific nodes are learned automatically by the architecture search.
\end{compactitem}

\vspace{1mm}
\noindent \textbf{Architecture search space.} The proposed MPNAS is developed based on the MobileNetV3-like search space within the TuNAS \cite{bender2020can} framework. 
The MobileNetV3-like~\cite{howard2019searching} search space represents the state-of-the-art mobile-friendly classification model. 
It consists of a stack of blocks with convolution layers, inverted bottleneck layers, hard-swish activation layers, a compact head, and optional Squeeze-and-Excite layers.
The searchable parameters includes the number of layers per block, the positions of Squeeze-and-Excite layers, kernel size, bottleneck expansion factors within $\{1,2, \cdots, 6\}$ and different filter sizes. 
For fair comparisons, we apply the TRL method on the MobileNetV3-like baseline model.

\vspace{1mm}
\noindent \textbf{Implementation details.} 
All the model search and training jobs are conducted on Cloud TPU V3 with 32 TPU chips.
%
Each search takes 90 epochs on ImageNet72. 
The batch size in search is 1024 for each task. 
The super-network model and the searched model weights are trained using RMSProp with a cosine decay learning rate schedule.
For search, the learning rate schedule starts at 8e-3. After it converges, the selected multi-path networks are trained from scratch with the learning rate schedule starting at 0.0325.
During search, our RL controllers start training after 1/8 total steps using Adam with a constant learning rate of 0.165. For the RL reward function~(\ref{eq:reward-function}), we set $T_0 = 84ms$ following \cite{bender2020can}. We tuned $\beta \in [-0.09, -0.03]$ and selected $\beta = -0.07$ based on the average accuracy.
The training stage runs until reaching 360 epochs on ImageNet72. 
The training stage batch size is set to 416 for each domain. 
%

\subsection{Model Accuracy}

Table~\ref{table:multihead-fulltrain} summarizes the top-1 accuracy for all the evaluated methods on the Visual Decathlon dataset. 
Compared to the SD-NAS approach, the performance of SP-MN degrades significantly due to the negative interference among domains.
The MDSP-NAS model slightly improves over the SP-MN model due to joint architecture search over all domains instead of searching on ImageNet72 only. 
The multi-path models, including MRP and MPNAS significantly improves the accuracy over the MDSP-NAS model by $25\%$ to $30\%$ due to the domain specific feature learning supported by the diverse path for each domain. 
MPNAS also significantly outperforms TRL on the MobileNetV3-Large architecture by $16.9\%$ since the MPNAS method supports selecting different operations between different paths if the two domains learn not to share. 
In addition, the proposed MPNAS model achieves a slight improvement of $1.9\%$ over the SD-NAS scheme because of the positive knowledge transfer among domains. 
The MPNAS method therefore achieves the best performance over all the MobileNetV3-like architectures.



\setlength{\tabcolsep}{4pt}
\begin{table}[tp]
\begin{center}
\caption{Top 1 accuracy (\%) of the highly correlated domains. Blue number indicates the best in each column.}
\label{table:accuracy-correlated}
\begin{tabular}{ccccc}
\hline\noalign{\smallskip}
Models & C100 & ImageNet72 & Flwr & Avg \\
\noalign{\smallskip}
\hline
\noalign{\smallskip}
SD-NAS & 69.63 & \textbf{\textcolor{blue}{61.87}} & 66.73 & 66.07 \\
MPNAS & \textbf{\textcolor{blue}{79.74}} & 59.62 & \textbf{\textcolor{blue}{71.76}} & \textbf{\textcolor{blue}{70.37}} \\
\hline
\end{tabular}
\end{center}
\end{table}
\setlength{\tabcolsep}{1.4pt}

To demonstrate the positive transfer effect among correlated domains, we evaluate the proposed MPNAS method on three domains which are visually similar and empirically correlated: ImageNet72, Cifar-100, and VGG\_Flowers. 
Table \ref{table:accuracy-correlated} shows that the MPNAS model can improve the accuracy by $10\%$ on Cifar-100 and $5\%$ on VGG\_Flower with very small ($2\%$) sacrifices on ImageNet72.

\subsection{Architecture Analysis}
\label{section:architecture-analysis}

\begin{figure}[tp]
\centering
\includegraphics[width=90mm]{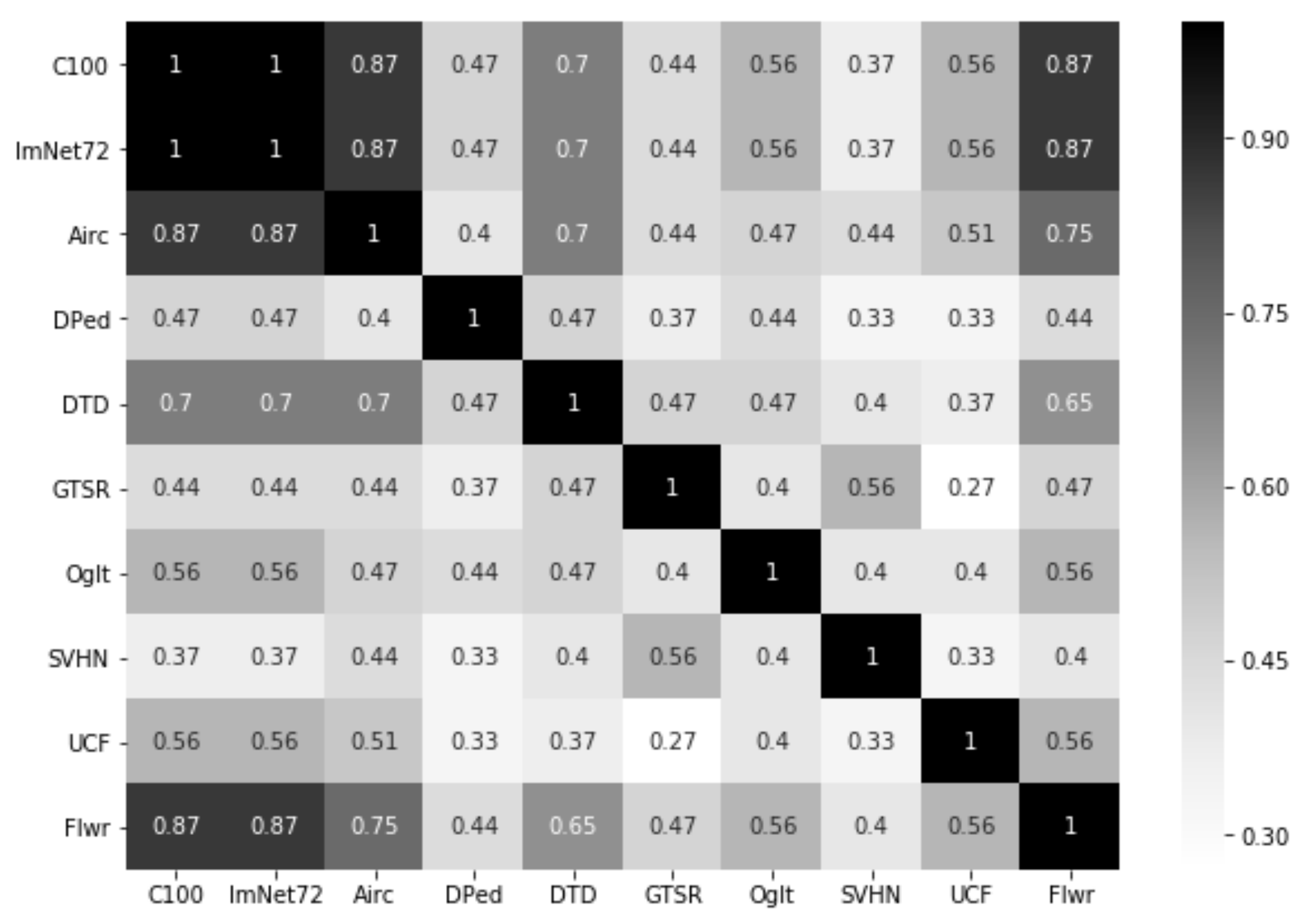}
\setlength{\abovecaptionskip}{0.cm}
\setlength{\belowcaptionskip}{-0.cm}
\caption{Confusion matrix for the Jaccard similarity score between the paths for the ten domains. 
The score value ranges from 0 to 1. 
A greater value indicates two paths share more nodes.}
\label{fig:jaccard-matrix}
\end{figure}

Intuitively, more correlated domains should share more nodes in order to maximize positive knowledge transfer. 
%
%
To measure architecture similarity between domains, we compute the pairwise Jaccard similarity score \cite{rajaraman2011mining} between the set of selected nodes in two domains, as shown in Fig.~\ref{fig:jaccard-matrix}. 
%
%
A higher Jaccard similarity score means the paths are more similar and thus the domains correlation is higher. 
The results show the paths generated by MPNAS are aligned with the empirical experience. 
%
For example, the paths for similar domains, e.g., ImageNet, Cifar-100 and VGG\_Flower, have high Jaccard similarity scores while the paths for dissimilar domains, e.g., GTSR and UCF, DPed and UCF, have low Jaccard similarity scores. 
These results again demonstrate that the proposed MPNAS can learn an effective parameter sharing strategy to increase positive transfer. 

\setlength{\tabcolsep}{4pt}
\begin{table*}[tp]
\begin{center}
\caption{Top 1 accuracy (\%) between the approaches using and not using adaptive balanced task prioritization. Blue numbers denote better results in pairwise comparisons.}
\label{table:domain-prioritization}
\begin{tabular}{c|cc|cc|cc|cc}
\hline\noalign{\smallskip}
\multirow{3}{*}{ Domains } & MDSP & MDSP & SP & SP & MRP & MRP & MPNAS & MPNAS \\
 & -NAS & -NAS & -MN & -MN &  &  &  &  \\
 &  & +ABDP & & +ABDP & & +ABDP & & +ABDP \\
\noalign{\smallskip}
\hline
\noalign{\smallskip}
ImageNet72 & 27.50 & \textbf{\textcolor{blue}{34.57}} & 21.84 & \textbf{\textcolor{blue}{30.15}} & 44.46 & \textbf{\textcolor{blue}{50.70}} & 48.46 & \textbf{\textcolor{blue}{57.02}} \\
Airc & 7.10 & \textbf{\textcolor{blue}{18.78}} & 5.29 & \textbf{\textcolor{blue}{6.19}} & 21.48 & \textbf{\textcolor{blue}{21.53}} & 26.12 & \textbf{\textcolor{blue}{29.65}} \\
C100 & 52.90 & \textbf{\textcolor{blue}{56.43}} & 50.26 & \textbf{\textcolor{blue}{59.45}} & \textbf{\textcolor{blue}{67.70}} & 67.35 & 74.86 & \textbf{\textcolor{blue}{81.16}} \\
DPed & 73.60 & \textbf{\textcolor{blue}{75.07}} & 51.00 & \textbf{\textcolor{blue}{59.87}} & 96.10 & 96.10 & \textbf{\textcolor{blue}{97.27}} & 96.02 \\
DTD & 32.40 & \textbf{\textcolor{blue}{30.80}} & 25.92 & \textbf{\textcolor{blue}{31.03}} & 30.71 & \textbf{\textcolor{blue}{31.49}} & 36.43 & \textbf{\textcolor{blue}{36.61}} \\
GTSR & 51.90 & \textbf{\textcolor{blue}{88.64}} & 40.77 & \textbf{\textcolor{blue}{52.50}} & \textbf{\textcolor{blue}{99.89}} & 99.77 & \textbf{\textcolor{blue}{99.92}} & 99.80 \\
Flwr & 59.90 & \textbf{\textcolor{blue}{65.59}} & 54.00 & \textbf{\textcolor{blue}{60.19}} & 60.96 & \textbf{\textcolor{blue}{61.75}} & 72.70 & \textbf{\textcolor{blue}{79.18}} \\
Oglt & 1.80 & \textbf{\textcolor{blue}{9.33}} & \textbf{\textcolor{blue}{1.59}} & 1.27 & \textbf{\textcolor{blue}{76.06}} & 75.76 & \textbf{\textcolor{blue}{76.34}} & 74.19 \\
SVHN & \textbf{\textcolor{blue}{27.60}} & 25.52 & 20.66 & \textbf{\textcolor{blue}{23.06}} & \textbf{\textcolor{blue}{92.29}} & 91.89 & \textbf{\textcolor{blue}{92.35}} & 92.34 \\
UCF & 27.00 & \textbf{\textcolor{blue}{47.53}} & 19.52 & \textbf{\textcolor{blue}{28.75}} & \textbf{\textcolor{blue}{71.11}} & 69.48 & \textbf{\textcolor{blue}{75.17}} & 71.79 \\
\hline
Avg & 36.17 & \textbf{\textcolor{blue}{45.23}} & 29.08 & \textbf{\textcolor{blue}{35.25}} & 66.08 & \textbf{\textcolor{blue}{66.58}} & 69.96 & \textbf{\textcolor{blue}{71.78}} \\ 
\hline
\end{tabular}
\end{center}
\end{table*}
\setlength{\tabcolsep}{1.4pt}

\setlength{\tabcolsep}{4pt}
\begin{table*}[hbt!]
\begin{center}
\caption{Top 1 accuracy (\%) achieved by various loss functions. Blue numbers are the best in the column.}
\label{table:task-prioritization-function}
\begin{tabular}{ccccccccccc|c}
\hline\noalign{\smallskip}
Funcs. & ImageNet72 & Airc. & C100 & DPed & DTD & GTSR & Flwr & OGlt & SVHN & UCF & Mean \\
\noalign{\smallskip}
\hline
\noalign{\smallskip}
UWL~\cite{kendall2018multi} & 41.78 & 24.65 & 72.09 & 95.40 & 34.46 & \textbf{\textcolor{blue}{99.93}} & 71.72 & 74.91 & 92.16 & \textbf{\textcolor{blue}{75.46}} & 68.26 \\
Empirical & 53.78 & 26.57 & 76.22 & 96.48 & 36.86 & 99.90 & 76.24 & 75.80 & 92.06 & 69.88 & 70.38 \\
$x$ & 48.46 & 26.12 & 74.86 & 97.27 & 36.43 & 99.92 & 72.70 & 76.34 & 92.35 & 75.17 & 69.96 \\
$x^2$ & 54.86 & 27.88 & 80.13 & 97.18 & 38.55 & 99.88 & 74.32 & 74.55 & 92.35 & 72.10 & 71.18 \\
$\exp(x/w)$ & 55.29 & 30.04 & 76.78 & 96.73 & 37.54 & 99.59 & 71.54 & 76.07 & 91.49 & 69.98 & 70.51 \\
$\exp(x/w\uparrow)$ & 49.16 & \textbf{\textcolor{blue}{31.17}} & 78.65 & \textbf{\textcolor{blue}{97.38}} & \textbf{\textcolor{blue}{38.89}} & 99.87 & 78.19 & \textbf{\textcolor{blue}{77.87}} & \textbf{\textcolor{blue}{92.49}} & 73.40 & 71.71 \\
$\exp(x/w\downarrow)$ & \textbf{\textcolor{blue}{57.02}} & 29.65 & \textbf{\textcolor{blue}{81.16}} & 96.02 & 36.61 & 99.80 & \textbf{\textcolor{blue}{79.18}} & 74.19 & 92.34 & 71.79 & \textbf{\textcolor{blue}{71.78}} \\ 
\hline
\end{tabular}
\end{center}
\end{table*}
\setlength{\tabcolsep}{1.4pt}

\subsection{Ablation Studies}

\label{sec:ablation_studies:ABDP}

\subsubsection{Adaptive Balanced Domain Prioritization} 
We carry out experiments on the MDSP-NAS, SP-MN, MRP, and MPNAS models with or without the proposed ABDP method for ablation studies. 
Table \ref{table:domain-prioritization} summarizes the pairwise experimental results.
Overall, applying the ABDP can always improve average training accuracy on Visual Decathlon dataset regardless of model architecture. 
For the single-path models (MDSP-NAS and SP-MN), the ABDP method greatly improves the performance since the model optimization on the nodes affects all the domains.
For the multi-path models (MRP and MPNAS), the performance gain is smaller 
since the joint loss gradient on a certain node does not affect domains that do not use this node. 
%
The results also show that the ADBP method is effective in prioritizing more difficult domains and achieving larger performance gains.

\subsubsection{Evaluation of Joint Optimizer} In this ablation study, we evaluate model accuracies with various MTL loss functions, including uncertainty weighted loss (UWL) \cite{kendall2018multi}, empirical weighted loss (Empirical), identity (no boosting, $x$), ABDP with quadratic polynomial boosting ($x^2$) and ABDP with the exponential boosting ($\exp(x/w)$) functions.

The exponential boosting loss function can be further enhanced by decreasing or increasing the domain prioritization coefficient $w$ during the training stage. 
For the linear decreasing schedule $\exp(x/w\downarrow)$, $w$ is decreased from $w_{max}$ to $w_{min}$ throughout the training process which means harder domains are weighted more in the later search/training stage (we use $w_{max}=2$ and $w_{min}=1$). 
For a linear increasing schedule $\exp(x/w\uparrow)$, $w$ is increased which means favoring harder domains in earlier stages. For the empirical weighted loss, we double the loss weight for ImageNet72 since it is one of the most challenging domains.

The results in Table \ref{table:task-prioritization-function} show that all the boosting functions outperform the uncertainty weighted loss (UWL) \cite{kendall2018multi} method. 
The exponential boosting function with decay coefficient schedule improves the average accuracy over UWL by $3.5\%$. 
That improvement also validates the previous assumption that transforming the loss subspace can help to optimize the model within the given training strategy over the weighted sum approaches. 
Moreover, both the exponential and square functions outperform the identity function which has no loss boosting. 
These results show that the proposed ABDP method improves MDL optimization when training jointly with multiple diverse domains. 
The linear decay schedule slightly improves the accuracy over the linear increasing schedule on the Visual Decathlon dataset as it reduces the interference between domains when the training process is converging. 

Compared with setting domain weights empirically, the ABDP approach does not require prior knowledge of relative domain difficulties, which makes it easier to generalize and adapt to different settings.

\section{Conclusions}

%
In this paper, we propose a multi-path NAS approach as the first work of applying efficient NAS to solve the challenges of on-device multi-domain learning (MDL), including
data imbalance, domain diversity, negative transfer, domain scalability, and
large search space of possible parameter sharing strategies.
By learning to selectively share parameters among domains, the multi-path network is able to reduce the total number of parameters and FLOPS, encourage positive knowledge transfer, and mitigate negative interference. 
The adaptive balanced domain prioritization algorithm achieves balanced performance among domains by dynamically adjusting the weighting of each domain during the search and training phases. 
Extensive experiments demonstrate that the proposed multi-path network achieves state-of-the-art accuracy on the 10-domain Visual Decathlon dataset with the mobile-friendly MobileNetV3-like network architecture. 
The selected model searched in the MobileNetV3-like search space reduces the number of parameters and FLOPS by $78\%$ and $32\%$, respectively compared to the approach by simply constructing the single-domain models without sacrificing model accuracy.

\clearpage

{\small
\bibliographystyle{ieee_fullname}
\bibliography{egbib}
}

\end{document}